\documentclass[letterpaper, 10 pt, journal, twoside]{IEEEtran}

\IEEEoverridecommandlockouts                              

\title{
GR-MG: Leveraging Partially-Annotated Data via Multi-Modal Goal-Conditioned Policy
}

\author{Peiyan Li$^{1,2}$$^{\dagger}$, Hongtao Wu$^{3*}$$^{\ddagger}$, Yan Huang$^{1,2*}$, Chilam Cheang$^{3}$,  Liang Wang$^{1,2}$, Tao Kong$^{3}$ 
\thanks{Manuscript received: August, 12, 2024; Revised November, 15, 2024; Accepted Decemeber, 17, 2024.}
\thanks{This paper was recommended for publication by Editor Markus Vincze upon evaluation of the Associate Editor and Reviewers' comments.}
\thanks{Digital Object Identifier (DOI): see top of this page.}
\thanks{$\dagger$ Work done during internship at Bytedance Research.}
\thanks{* Corresponding author  $\ddagger$ Project lead}
\thanks{$^{1}$New Laboratory of Pattern
Recognition (NLPR), State Key Laboratory of Multimodal Artificial Intelligence Systems (MAIS), Institute of
Automation, Chinese Academy of Sciences. $^{2}$School of Artificial Intelligence, University of Chinese Academy of Sciences. $^{3}$ByteDance Research} 
\thanks{{\tt\small peiyan.li@cripac.ia.ac.cn}}
\thanks{{\tt\small wuhongtao.123@bytedance.com}}
\thanks{{\tt\small yhuang@nlpr.ia.ac.cn}}
}

\usepackage{cite}
\usepackage[T1]{fontenc}
\usepackage{amsmath}
\usepackage{graphicx} 
\usepackage{multirow}
\usepackage{threeparttable}
\usepackage{url}
\usepackage{amssymb}
\usepackage{bbding}
\usepackage{xcolor}
\usepackage{booktabs}
\usepackage{balance}
\usepackage{float}
\usepackage{multicol}
\usepackage{diagbox}
\raggedcolumns 
\begin{document}

\maketitle

\begin{figure*}[!t]
\centering
\includegraphics[width=1.8\columnwidth]{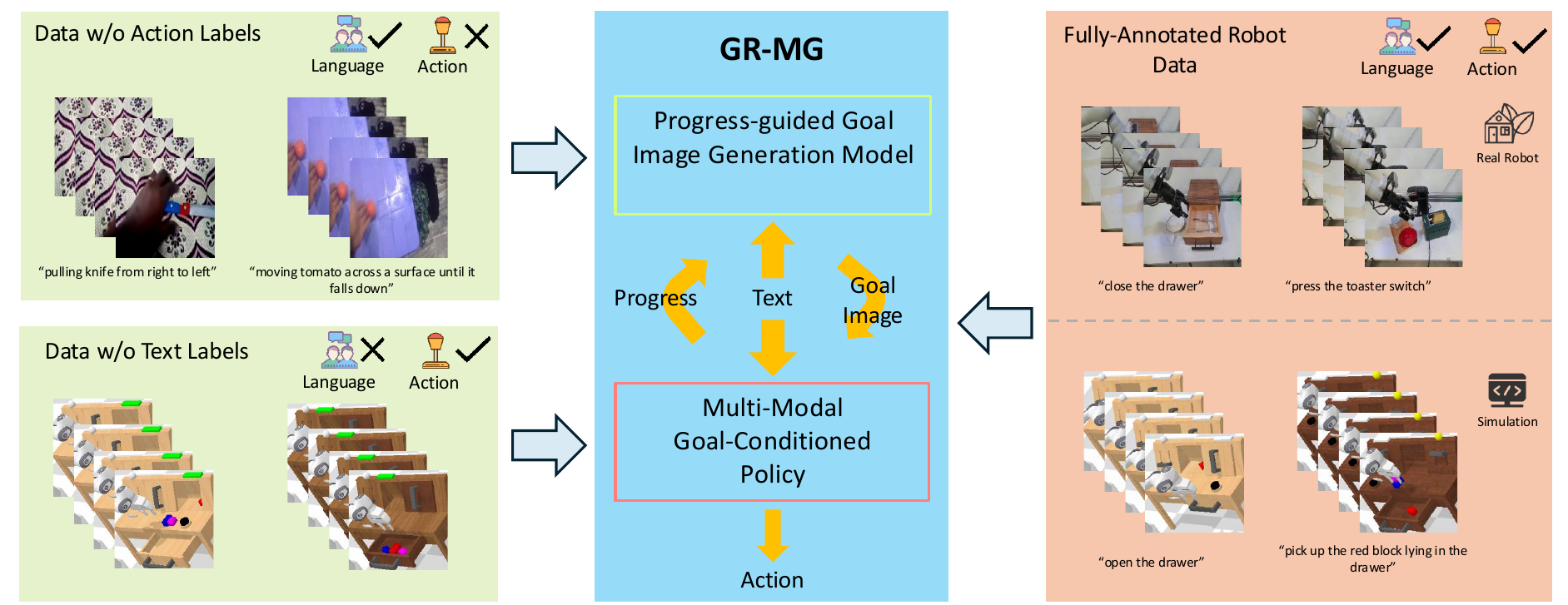}
\caption{
\textbf{Overview.}
GR-MG consists of two modules: a progress-guided goal image generation model and a multi-modal goal-conditioned policy. The former generates a goal image based on the current observation, text instruction, and task progress. The latter predicts task progress and actions based on the text and the goal image produced by the former.
Data without action labels can be used to train the goal image generation model, while the multi-modal goal-conditioned policy can leverage data without text labels. 
Fully-annotated data is used for training both modules.}
\label{fig:overview}
\vspace{-0.5cm}
\end{figure*}
\begin{abstract}
The robotics community has consistently aimed to achieve generalizable robot manipulation with flexible natural language instructions. 
One primary challenge is that obtaining robot trajectories fully annotated with both actions and texts is time-consuming and labor-intensive. 
However, partially-annotated data, such as human activity videos without action labels and robot trajectories without text labels, are much easier to collect. 
Can we leverage these data to enhance the generalization capabilities of robots? 
In this paper, we propose GR-MG, a novel method which supports conditioning on a text instruction and a goal image.
During training, GR-MG samples goal images from trajectories and conditions on both the text and the goal image or solely on the image when text is not available.
During inference, where only the text is provided, GR-MG generates the goal image via a diffusion-based image-editing model and conditions on both the text and the generated image.
This approach enables GR-MG to leverage large amounts of partially-annotated data while still using languages to flexibly specify tasks.
To generate accurate goal images, we propose a novel progress-guided goal image generation model which injects task progress information into the generation process.
In simulation experiments, GR-MG improves the average number of tasks completed in a row of 5 from 3.35 to 4.04.
In real-robot experiments, GR-MG is able to perform 58 different tasks and improves the success rate from 68.7\% to 78.1\% and 44.4\% to 60.6\% in simple and generalization settings, respectively.
It also outperforms comparing baseline methods in few-shot learning of novel skills.
Video demos, code, and checkpoints are available on the project page: \url{https://gr-mg.github.io/}.
\end{abstract}

\begin{IEEEkeywords}
Deep Learning in Grasping and Manipulation, Imitation Learning, Learning from Demonstration
\end{IEEEkeywords}

\section{INTRODUCTION}
\label{sec:intro}
\IEEEPARstart{T}{he} robotics research community is striving to achieve generalizable robot manipulation using language-based instructions. 
Among various methods, imitation learning from human demonstrations is one of the most promising endeavors\cite{brohan2022rt, brohan2023rt, team2024octo, wu2023unleashing, li2023vision}.
However, human demonstration data are scarce.
And the process of collecting human demonstrations with action \textit{and} text labels is time-consuming and labor-intensive.
On the other hand, trajectories without language annotations are a scalable source of data. 
They do not require executing pre-defined tasks or text labeling, and can be collected without human constant supervision~\cite{ahn2024autort}.
These data can be automatically labeled with hindsight goal relabeling \cite{lynch2021language}: any frame in a video can be used as a goal image to condition the policy for predicting actions to evolve towards this frame from previous frames.
Furthermore, there is a substantial collection of text-annotated human activity videos without action labels from various public datasets~\cite{goyal2017something, miech2019howto100m, grauman2022ego4d}.
These data contain valuable information about how the agent should move to change the environment according to the text description.
Can we develop a policy to effectively leverage all the above partially-annotated data?

Previous methods have ventured into this domain, but most were limited to using data without either text labels or action labels \cite{wu2023unleashing, lynch2021language, mendonca2023structured, baker2022video, mees2023grounding}. 
Recent initiatives introduce diffusion models for generating goal images \cite{black2023zero} and future videos \cite{du2024learning, du2023video}. 
The generated image or video is then used as inputs for goal-conditioned policies or inverse dynamics models to predict actions, enabling the use of all the above-mentioned partially-annotated data.
However, these approaches often overlook crucial information, such as task progress, during the goal generation phase. 
This can lead to inaccurate generated goals which may significantly affect the subsequent action prediction.
Furthermore, these policies rely solely on the goal image or video for action prediction, making them brittle in the case where the generated goal is inaccurate.

To tackle these issues, we introduce GR-MG, a model designed to support multi-modal goals.
It comprises two modules: a progress-guided goal image generation model and a multi-modal goal-conditioned policy.
GR-MG is able to leverage data without action labels (\textit{e.g.}, text-annotated human activity videos) to train the goal image generation model along with the fully-annotated robot trajectories.
And data without text labels can be used for training the multi-modal goal-conditioned policy.
Given that robot manipulation is a sequential decision-making process, we incorporate a novel task progress condition into the goal image generation model.
This significantly improves the performance on both the goal image generation and action prediction.
During training, we sample goal images from trajectories and condition the policy on both the goal image and the text instruction or solely the goal image if text is unavailable.
During inference, the policy utilizes both the text instruction and the goal image generated by the goal image generation model to predict actions.
Since GR-MG is conditioned on both the text and goal image, the policy can still rely on the text condition to guide action prediction even if the generated goal image is inaccurate, substantially improving the robustness of the model.

We perform extensive experiments on the challenging CALVIN simulation benchmark~\cite{mees2022calvin} and a real-robot platform.
In CALVIN, without using additional partially-annotated data, GR-MG significantly outperforms all the comparing state-of-the-art methods in a zero-shot generalization setting.
It improves the success rates of completing 1 and 5 tasks in a row from 93.8\% to 96.8\% and 41.2\% to 64.4\%, respectively.
When incorporating data without text labels into training, GR-MG achieves an average length of 3.11 with only 10\% of the provided fully-annotated data, outperforming the competitive GR-1~\cite{wu2023unleashing} baseline which uses all the data.
In real-robot experiments, we evaluate GR-MG in a simple and four challenging generalization settings.
In total, GR-MG is able to perform 58 tasks and improves the average success rate from 68.7\% to 78.1\% in the simple setting and from 44.4\% to 60.6\% in generalization settings.
GR-MG also surpasses comparing baseline methods in few-shot learning of novel skills.
In summary, the contribution of this paper is threefold:
\begin{itemize}
    \item We propose a novel \textbf{G}enerative \textbf{R}obot Policy with \textbf{M}ulti-modal \textbf{G}oals (GR-MG), which is able to leverage both data without action labels and data without text labels during training.
    \item We introduce the task progress condition in goal image generation, substantially improving the accuracy of the generated goal image.
    \item We perform extensive experiments and ablation studies in simulation and the real world to verify the effectiveness of GR-MG in simple, generalization, and few-shot learning settings.
\end{itemize}

\begin{figure*}[!t]
\centering
\includegraphics[width=6in]{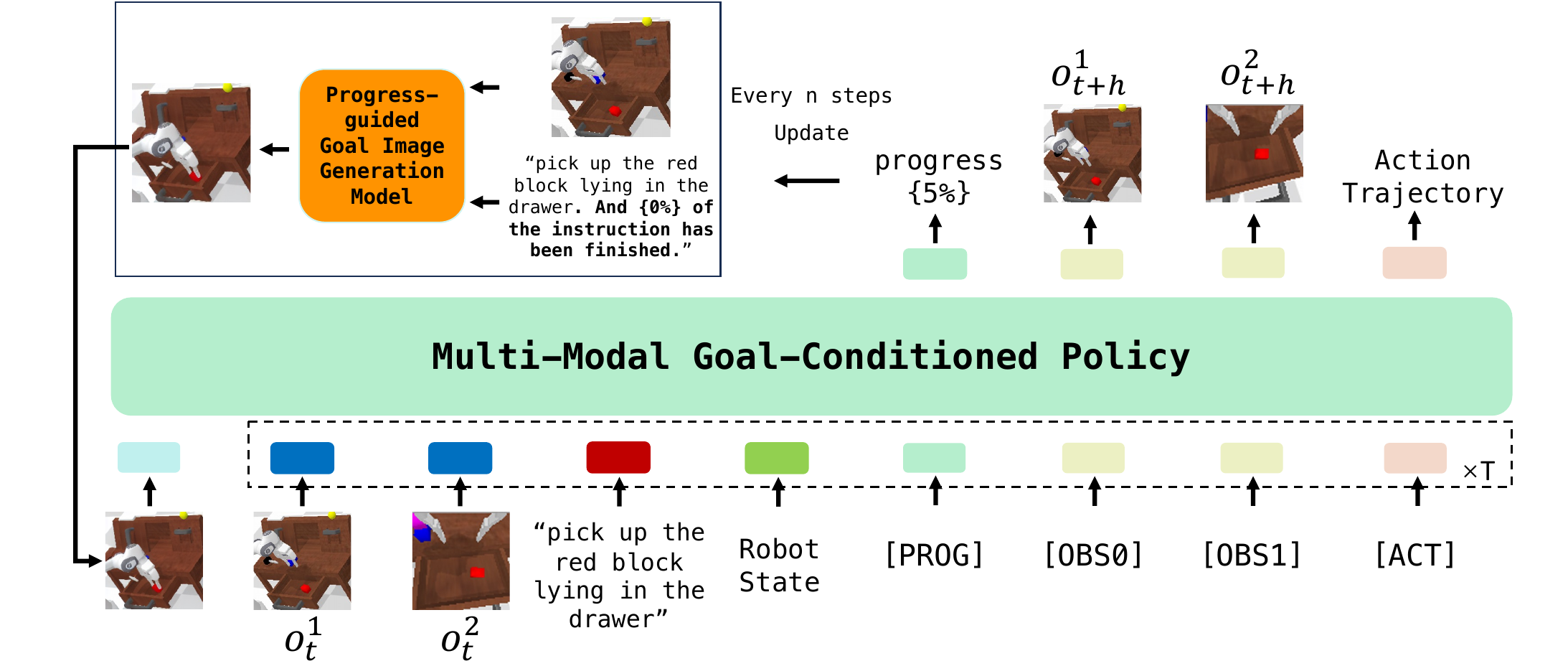}
\caption{
\textbf{Network Architecture.}
We use a diffusion model to generate the goal image based on the current image, text instruction, and task progress. 
The generated goal image is then sent to the multi-modal goal-conditioned policy, which takes as inputs the text instruction, sequences of observations images and robot states. 
The policy is a GPT-style transformer. 
We insert query tokens \texttt{[PROG]}, \texttt{[OBS]}, and \texttt{[ACT]} after the input tokens to predict actions, future images, and the task progress, respectively.
The progress is fed back to the goal image generation model. 
In this figure, we show the setup with two camera views: a static view and a view captured from a wrist-mounted camera.
} 
\label{fig:architecture}
\vspace{-0.3cm}
\end{figure*}

\section{Related Work}
\label{sec:related_work}
\subsection{Leveraging Multi-Modal Goals in Policy Learning}
Language is probably the most flexible and intuitive way for a human to specify tasks for a robot~\cite{brohan2022rt, brohan2023rt, wu2023unleashing, li2023vision, bharadhwaj2023roboagent}.
However, language and visual signals exist in different domains.
And the information contained in a language instruction may be abstract for a visual policy to comprehend.
A line of research explores goal-image conditioned policies~\cite{bousmalis2023robocat,shah2023vint}.
But obtaining goal images during rollouts is challenging in robot manipulation tasks.
Recently, some studies have proposed leveraging both language and visual goals (goal images or videos) as conditions during training~\cite{jang2022bc, lynch2021language, reuss2024multimodal, mees2022matters,jiang2022vima,team2024octo}.
And Mutex~\cite{shah2023mutex} is able to leverage multiple modalities for task specification, including speech, text, images, and videos.
However, in language-conditioned manipulation, where language is the only condition provided, these methods do not generate goal images. Instead, they rely solely on the language to condition the prediction of actions during inference.
GR-MG differs from these methods in that it uses both the language \textit{and} the goal image during training and inference.
And the goal image is generated via a diffusion model from the language and current observation.
The most related work to GR-MG is Susie~\cite{black2023zero}.
It also uses a diffusion model to generate the goal image.
The goal image is then fed into a diffusion policy~\cite{chi2023diffusion} for action prediction.
GR-MG follows a similar paradigm but differs in two key aspects. 
First, it introduces a novel task progress condition in goal image generation, significantly enhancing the accuracy of the generated goal image. 
Second, GR-MG conditions the policy with both the language and goal image, whereas Susie relies solely on the generated goal images. 
This design enables GR-MG to address situations where the generated goal image is inaccurate, thereby improving its robustness in challenging out-of-distribution settings.

\subsection{Leveraging Partially-Annotated Data in Policy Learning}
Given that collecting fully-annotated robot data with action and language labels is time-consuming, many studies turn to alternative sources of data with partial annotations. 
One approach involves learning useful representations from text-annotated videos without action labels, which are subsequently used for downstream policy learning~\cite{nair2022r3m, xiao2022masked, majumdar2023we}.
Recently, RT-2~\cite{brohan2023rt} introduces a vision-language-action model that incorporates both robot data and large-scale vision-language data during training, demonstrating powerful generalization capabilities.
Another popular method is to learn a video model or world model from existing video data to predict future videos~\cite{yang2023learning, du2023video, du2024learning, mendonca2023structured}. 
An inverse dynamics model or goal-conditioned policy is then employed to predict actions based on the predicted video. 
Recently, Wu et al.~\cite{wu2023unleashing} proposes GR-1, which is pretrained on large-scale text-annotated videos and then finetuned on robot trajectories with text labels.
GR-MG differs from GR-1 in two aspects. 
First, GR-MG leverages data without text labels during training, while GR-1 does not. 
Second, GR-MG utilizes both the language and goal image as conditions, whereas GR-1 relies solely on the language. 
These differences significantly enhance GR-MG's generalization capabilities.

\section{Method}
\label{sec:method}
We aim to learn a policy $\pi$ for language-conditioned visual robot manipulation. 
Specifically, the policy takes as inputs a text instruction $l$, a sequence of observation images $\mathbf{o}_{t-h:t}$, and a sequence of robot states $\mathbf{s}_{t-h:t}$.
The observation images $\mathbf{o}_{t-h:t}$ include RGB images captured from timestep $t-h$ to $t$.
In this paper, we use a static camera and a wrist-mounted camera to capture these images.
The robot states $\mathbf{s}_{t-h:t}$ include the 6-DoF poses of the end-effector and the binary gripper states from timestep $t-h$ to $t$.
The policy outputs an action trajectory $\mathbf{a}_{t}$ in an end-to-end manner:
\begin{equation}
    \mathbf{a}_{t} = \pi(l, \mathbf{o}_{t-h:t}, \mathbf{s}_{t-h:t})
\end{equation}

\subsection{Data}
Training a language-conditioned policy typically requires fully-annotated robot trajectories $\tau$ which contain both action and text labels:
\begin{equation}
    \tau = \{ l, (\mathbf{o}_{1}, \mathbf{s}_{1}, \mathbf{a}_{1}), (\mathbf{o}_{2}, \mathbf{s}_{2}, \mathbf{a}_{2}), \ldots, (\mathbf{o}_{T}, \mathbf{s}_{T}, \mathbf{a}_{T}) \} 
\end{equation}
However, collecting fully-annotated data is time-consuming. 
The motivation behind GR-MG is to leverage partially-annotated data in training.
There are two main types of partially-annotated data: data without action labels and data without text labels. 
For brevity, we abbreviate them as data w/o action labels and data w/o text labels, respectively, throughout the rest of the paper.

Data w/o action labels contains videos with text annotations but does not include action labels.
This type of data can be readily sourced from various public datasets~\cite{goyal2017something, miech2019howto100m, grauman2022ego4d}. Data w/o text labels, on the other hand, contains videos with corresponding action labels but does not include text annotations. 
They can be collected on a large scale by allowing robots to execute policies without constant human supervision~\cite{ahn2024autort}.
GR-MG is designed to leverage both types of partially-annotated data along with fully-annotated data to maximum the usage of data available.

\subsection{Network Architecture}
The network architecture of GR-MG is illustrated in Fig.~\ref{fig:architecture}.
It consists of two modules: a progress-guided goal image generation model and a multi-modal goal-conditioned policy.

\subsubsection{Progress-guided Goal Image Generation Model}
This module generates the goal image based on the current observation image, a text description of the task, and the task progress. 
Specifically, we use InstructPix2Pix~\cite{brooks2023instructpix2pix}, a diffusion-based image-editing model, as the network. 
Following the approach in Susie~\cite{black2023zero}, we generate a sub-goal representing an intermediate state rather than the final state, and update the sub-goal at fixed time intervals during the rollout.
However, unlike Susie which generates the goal image solely based on the current observation and the text instruction, we incorporate task progress information in the image generation process.
The insight is that robot manipulation is a sequential decision process instead of static image editing.
Additionally, the current observation can be ambiguous in the case where there are identical observations throughout the rollout of a task, \textit{e.g.}, moving an object back-and-forth.
The progress information can provide valuable global information for goal image generation.
During training, the progress of a frame can be easily obtained from its timestep in the trajectory (or video).
During inference, such information is not accessible.
To address this, we train the policy to predict task progress alongside the action trajectory, as outlined in the following paragraph.
To incorporate progress information into the generation model, we append a sentence to the task description, \textit{e.g.}, "pick up the red block. \textbf{And 60\% of the instruction has been completed.}".
We use T5-Base~\cite{raffel2020exploring} to encode the text.

\subsubsection{Multi-modal Goal-Conditioned Policy}
Following GR-1~\cite{wu2023unleashing}, we employ a similar GPT-style transformer network~\cite{radford2019language}.
We briefly review GR-1 for completeness. 
GR-1 takes as inputs a text instruction, a sequence of observation images, and a sequence of robot states.
It outputs actions and future images.
The inputs are encoded into tokens with corresponding encoders~\cite{radford2021learning, he2022masked} or linear layers.
GR-1 leverages an \texttt{[ACT]} query token and multiple \texttt{[OBS]} query tokens for action and future image prediction, respectively.
The query tokens and the input tokens are concatenated together before being fed into the transformer.
The output embedding of the \texttt{[ACT]} token are passed through linear layers to predict actions.
The output embeddings of the \texttt{[OBS]} tokens are concatenated with learnable mask tokens to predict future images with a transformer.
We refer readers to~\cite{wu2023unleashing} for more details.

GR-MG differs from GR-1 in three ways.
Firstly, GR-MG is a multi-modal goal-conditioned policy which takes a goal image \textit{and} a text instruction as conditions; GR-1 only uses the text for conditioning.
To inject the goal image condition, we use MAE~\cite{he2022masked} to tokenize the goal image generated by the goal image generation model and append the tokens at the front of the input token sequence (Fig.~\ref{fig:architecture}).
These tokens can be attended by all the subsequent tokens for conditioning.
Secondly, in order to predict task progress, a \texttt{[PROG]} query token is included.
The output embedding of the token is passed through linear layers to regress the progress value.
Finally, we follow recent work~\cite{zhao2023learning} to predict action trajectories instead of a single action as in GR-1 via a conditional variational autoencoder (cVAE)~\cite{sohn2015learning, kingma2013auto}.
Specifically, we use a VAE encoder to encode the action trajectory into a style vector embedding.
We concatenate the style vector embedding, the output embedding of the $\texttt{[ACT]}$ token, and $k$ learnable token embeddings together and input them to a transformer for predicting an action trajectory of $k$ steps.

\subsection{Training}
To train the goal image generation model, we follow the training setting in InstructPix2Pix~\cite{brooks2023instructpix2pix} and train a noise prediction model as in DDPM~\cite{ho2020denoising}.
We sample the frame which is $N$ steps away from the current frame in the trajectory/video as the goal image.
As a rough task progress can already be very informative and can bring about more stable training, we discretize the task progress predicted from the policy from 0\% to 100\% into 10 bins before passing it to the goal image generation model.
As the training of goal image generation model only requires videos with text annotations, we are able to incorporate data w/o action labels in the training alongside the fully-annotated robot trajectories.

For training the multi-modal goal-conditioned policy, we initialize its weight with the pre-trained model weight derived from the generative video pre-training on Ego4d~\cite{grauman2022ego4d} in alignment with GR-1~\cite{wu2023unleashing}.
The multi-modal goal conditioning allows the policy to utilize partially-annotated data w/o text labels.
Specifically, if data w/o text labels is available, we first train the policy with the data, where a null string is provided as the text condition.
After that, we finetune the policy on fully-annotated robot trajectories.

The input task progress in the training of the goal image generation model is computed from the timestep of the image frame in the video.
The goal image in the training of the policy is sampled from the ground-truth goal image in the trajectory.
Therefore, the training of the two modules are independent.
We leave the investigation of jointly training the two modules as future work.

\subsection{Inference}
During inference, the task progress is initially set as zero.
And the goal image generation model uses the current observation image, text instruction, and task progress to generate the goal image.
The goal image is passed to the multi-modal goal-conditioned policy together with the text instruction, sequence of observation images, and sequence of robot states to predict action trajectories and task progress.
As the goal image is set as $N$ steps away from the current image in training, we run the goal image generation model only every $n < N$ steps in a closed-loop manner for efficiency.

\section{Experiments}
\label{sec:experiments}
We perform extensive experiments in a simulation benchmark and the real world to evaluate the performance of GR-MG.
We aim to answer three questions.
1) Can GR-MG perform multi-task learning and deal with challenging generalization settings including unseen distractors, unseen instructions, unseen backgrounds, and unseen objects?
2) Does incorporating data w/o text labels enhance the performance of GR-MG, especially when fully-annotated data is scarce?
3) How does the inclusion of data w/o action labels improve the performance of GR-MG?
4) Can GR-MG learn novel skills in a few-shot setting?
{
 \renewcommand{\arraystretch}{1.2}
\begin{table*}[]
\caption{Results On the ABC-\textgreater{}D Splits of CAVLIN Benchmark}
\centering
\resizebox{0.9\textwidth}{!}{
\begin{tabular*}{\textwidth}{@{\extracolsep{\fill}}ccccccccc}
\hline
\multirow{2}{*}{Method} & \multirow{2}{*}{\begin{tabular}[c]{@{}c@{}}Fully-Annotated \\ Data\end{tabular}} & \multirow{2}{*}{\begin{tabular}[c]{@{}c@{}}Partially-Annotated \\ Data\end{tabular}} & \multicolumn{6}{c}{No. Instructions in a Row (1000 chains)}                                                  \\ \cline{4-9} 
                        &                                                                                  &                            & 1               & 2               & 3               & 4               & 5               & Avg. Len.         \\ \hline
Hulc\cite{mees2022matters}                    & 100\%                                                                            & \checkmark                        & 41.8\%          & 16.5\%          & 5.7\%           & 1.9\%           & 1.1\%           & 0.67±0.10          \\
MDT\cite{reuss2024multimodal}                     & 100\%                                                                            & \checkmark                       & 61.7\%          & 40.6\%          & 23.8\%          & 14.7\%          & 8.7\%           & 1.54±0.04          \\
Spil\cite{zhou2023language}                    & 100\%                                                                            & \checkmark                       & 74.2\%          & 46.3\%          & 27.6\%          & 14.7\%          & 8.0\%           & 1.71               \\
Roboflamingo\cite{li2023vision}            & 100\%                                                                            & $\times$                         & 82.4\%          & 61.9\%          & 46.6\%          & 33.1\%          & 23.5\%          & 2.47               \\
Susie\cite{black2023zero}                   & 100\%                                                                            & \checkmark                       & 87.0\%          & 69.0\%          & 49.0\%          & 38.0\%          & 26.0\%          & 2.69               \\
GR-1\cite{wu2023unleashing}                    & 100\%                                                                            & $\times$                         & 85.4\%          & 71.2\%          & 59.6\%          & 49.7\%          & 40.1\%          & 3.06               \\
3D Diff Actor\cite{ke20243d}           & 100\%                                                                            & $\times$                          & 93.8\%          & 80.3\%          & 66.2\%          & 53.3\%          & 41.2\%          & 3.35±0.04          \\
GR-MG w/o image         & 100\%                                                                            & $\times$                          & 91.0\%          & 79.1\%          & 67.8\%          & 56.9\%          & 47.7\%          & 3.42±0.28          \\
GR-MG w/o text          & 100\%                                                                            & $\times$                         & 91.8\%          & 79.8\%          & 68.9\%          & 58.1\%          & 48.1\%          & 3.46±0.04          \\
GR-MG w/o progress          & 100\%                                                                            & $\times$                         & 94.1\%          & 85.0\%          & 75.2\%          & 65.4\%          & 56.3\%          & 3.76±0.11          \\
GR-MG (Ours)                  & 100\%                                                                            & $\times$                         & \textbf{96.8\%} & \textbf{89.3\%} & \textbf{81.5\%} & \textbf{72.7\%} & \textbf{64.4\%} & \textbf{4.04±0.03} \\ \hline
GR-1\cite{wu2023unleashing}                    & 10\%                                                                             & $\times$                         & 67.2\%          & 37.1\%          & 19.8\%          & 10.8\%          & 6.9\%           & 1.41±0.06          \\
GR-MG w/o part. ann. data                 & 10\%                                                                             & $\times$                         & 82.4\%          & 60.8\%          & 42.2\%          & 28.7\%          & 19.7\%          & 2.33±0.04          \\
GR-MG (Ours)                   & 10\%                                                                             & \checkmark                        & \textbf{90.3\%} & \textbf{74.5\%} & \textbf{61.2\%} & \textbf{47.4\%} & \textbf{37.5\%} & \textbf{3.11±0.08} \\ \hline
\end{tabular*}}
\label{tab:calvin}
\begin{tablenotes}
\footnotesize
\item 
"100\%" and "10\%" denote the proportion of fully-annotated data used for training. 
"w/o progress" indicates that task progress is not provided for the goal image generation model. 
"w/o text" indicates that the policy uses only the goal image as the condition.
"w/o image" indicates that the policy uses only the text as the condition.
"w/o part. ann. data" indicates that the policy does not use the trajectories without text labels for training.
\end{tablenotes}
\end{table*}

\begin{figure*}[!t]
\centering
\includegraphics[width=0.75\textwidth]{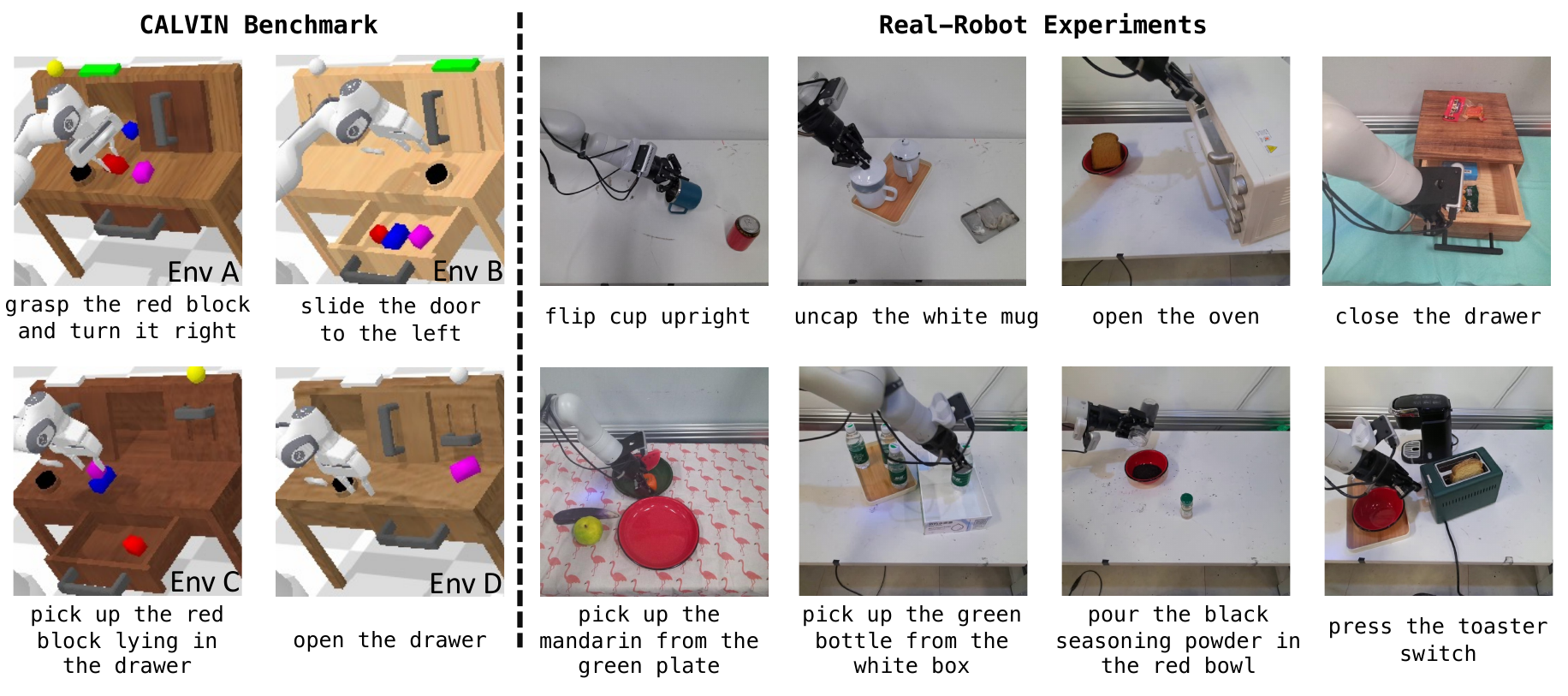}
\caption{\textbf{Experiments.} CALVIN benchmark consists of 34 tasks across four different environments. 
Real-robot experiments encompass 58 tasks, including pick-and-place and non-pick-and-place manipulations.
}
\label{fig:data}
\vspace{-0.3cm}
\end{figure*}

\subsection{CALVIN Benchmark Experiments}
\subsubsection{Experiment Settings}
We perform experiments in the challenging CALVIN~\cite{mees2022calvin} benchmark environment.
CALVIN focuses on language-conditioned visual robot manipulation.
It contains 34 tasks and four different environments, \textit{i.e.}, Env A, B, C, and D (Fig.~\ref{fig:data}).
Each environment contains a Franka Emika Panda robot with a parallel-jaw gripper and a desk for performing different manipulation tasks.

As indicated by~\cite{mees2022calvin}, only 1\% of the collected trajectories were annotated with texts.
The remaining trajectories without text labels contain unstructured play data collected by untrained users with no information about the downstream tasks.
These trajectories can be used as data w/o text labels for training the multi-modal goal-conditioned policy of GR-MG.
We refer the readers to~\cite{mees2022calvin} for more details on the dataset.

\subsubsection{Generalization to Unseen Environments}
We perform experiments on the ABC$\rightarrow$D split: we train GR-MG with data collected from Env A, B, and C and evaluate in Env D.
This experiment helps us evaluate the performance of GR-MG on handling unseen environments.
In total, the number of fully-annotated trajectories from Env A, B, and C is about 18k.
We use these data to train GR-MG.
Following the evaluation protocol from~\cite{mees2022calvin}, we command the robot to perform 1,000 sequences of tasks. 
Each sequence consists of five tasks in a row.
The language instruction for a new task is sent to the robot only after the current task is successfully completed.
We train GR-MG with three seeds and evaluate the performance of the last five epochs for each seed.
We report the mean and variance of all these checkpoints.

We compare GR-MG with various state-of-the-art methods. 
Results are shown in Tab.~\ref{tab:calvin}. 
Results of baseline methods are sourced from the original papers or provided by the authors.
GR-MG outperforms all the comparing baseline methods, establishing a new state-of-the-art. 
The average length reported in the last column of Tab.~\ref{tab:calvin} serves as a comprehensive metric, indicating the average number of tasks the robot is able to accomplish in a row of five across the evaluated 1,000 sequences. 
GR-MG improves the average length from 3.35 to 4.04.
Additionally, it increases the success rate of completing one and five tasks in a row from 93.8\% to 96.8\% and from 41.2\% to 64.4\%, respectively. 
These results showcase that GR-MG possesses strong capabilities of multi-task learning and generalization to unseen environments. 

\begin{figure*}[!t]
\centering
\includegraphics[width=0.85\textwidth]{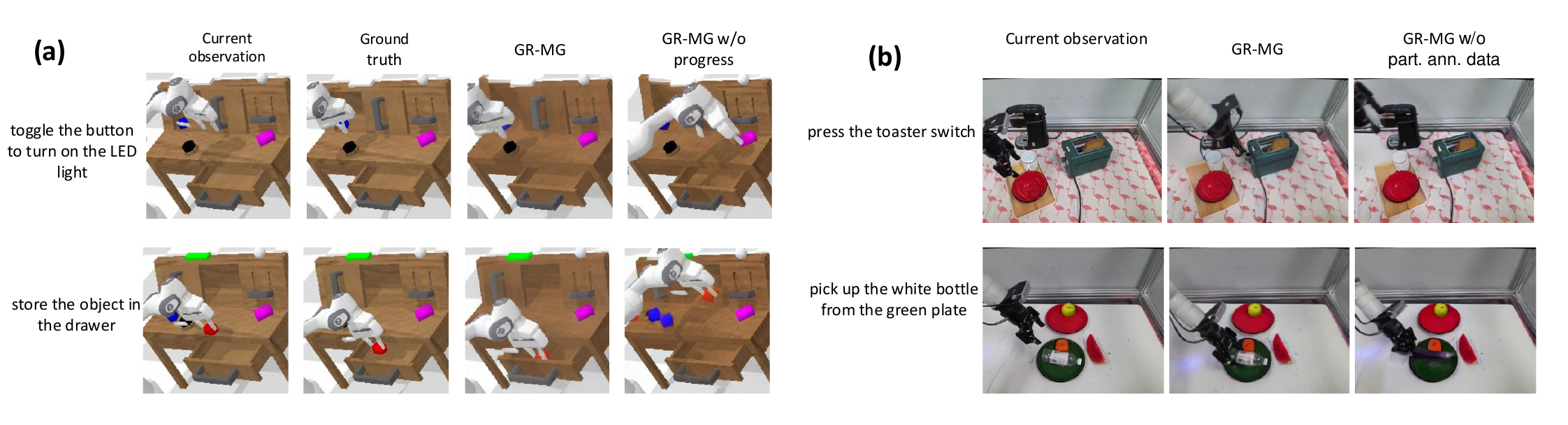}
\caption{\textbf{Visualization of the generated goal images in CALVIN Benchmark and Real-Robot Experiments.} 
(a) The generated images of GR-MG closely align with the ground truths. 
Without task progress information, the goal images generated by GR-MG w/o progress diverge from the text instructions.
(b) Without partially-annotated data, the generated goal images do not adhere to the text instructions and suffer from hallucination.
}
\label{fig:compare_ip2p}
\vspace{-0.3cm}
\end{figure*}

\subsubsection{Data Scarcity}
Obtaining large-scale fully-annotated trajectories is challenging, especially in the real world.
In this experiment, we simulate the data scarcity challenge by training GR-MG with only 10\% of the fully-annotated data in the ABC$\rightarrow$D split (about 1.8k trajectories for 34 tasks).
However, we make use of all the data without text labels in Env A, B, and C to train the policy first before finetuning on the 10\% fully-annotated trajectories.
The data w/o text labels contains 1M frames, while the 10\% fully-annotated data contains 0.1M frames.
Through this experiments, we hope to verify the effectiveness of incorporating data w/o text labels in enhancing the performance of policy learning.
Results are shown in Tab.~\ref{tab:calvin}.
GR-MG significantly outperforms GR-1~\cite{wu2023unleashing}, improving the success rate from 67.2\% to 90.3\% and the average length from 1.41 to 3.11. 
We also compare with a variant of our method, GR-MG w/o partially-annotated data (GR-MG w/o part. ann. data), which does not utilize the data w/o text labels during training. 
GR-MG significantly outperforms this variant.
Based on our observation, this variant is capable of generating correct goal images most of the time. 
However, the policy often struggles to accurately follow the generated goal image. 
This justifies the usefulness of data w/o text labels in training the multi-modal goal-conditioned policy, especially in the case where fully-annotated data is scarce.

\subsubsection{Ablation Studies}
To assess the impact of using multi-modal goals as conditions, we compare with two variants, GR-MG w/o text and GR-MG w/o image, which remove the text condition and goal image condition in the policy during training, respectively.
We retrain the two variants to ensure fair comparison.
The performance of these two variants is similar.
GR-MG significantly outperforms both of them, indicating that both modalities are essential for effective policy conditioning. 
To investigate the effectiveness of task progress, we also compare with GR-MG w/o progress, which excludes the task progress information in the goal image generation model. 
GR-MG surpasses this variant, showcasing the effectiveness of incorporating task progress information.

In addition, we perform quantitative comparison on the goal images generated by GR-MG and GR-MG w/o progress.
We compare the similarity between generated goal images and ground truths using four metrics: Mean Squared Error (MSE), Peak Signal-to-Noise Ratio (PSNR), Structural Similarity Index Measure (SSIM), and Cosine Distance of ResNet50 Features (CD-ResNet50). 
Results are shown in Tab.~\ref{tab:compare_ip2p_no_ip2p_progress}.
GR-MG outperforms GR-MG w/o progress in all four metrics, underscoring the effectiveness of incorporating task progress information in goal image generation.
Qualitative results are shown in Fig.~\ref{fig:compare_ip2p}(a).
Although GR-MG w/o progress is able to generate goal images with high visual quality, these images do not align well with the text instructions. 
In contrast, GR-MG accurately follows the text instructions and generates goal images that closely match the ground truths. 

\begin{table}[t]
\caption{Ablation Studies on Task Progress for Goal Image Generation}
\centering
\begin{tabular}{@{}cllcccc@{}}
\toprule
\multicolumn{3}{c}{Method}        & MSE ↓             & PSNR ↑           & SSIM ↑         & CD-ResNet50 ↑   \\ \midrule
\multicolumn{3}{c}{GR-MG w/o progress} & 965.347          & 18.821          & 0.721         & 0.945          \\
\multicolumn{3}{c}{GR-MG (Ours)}      & \textbf{903.139} & \textbf{19.121} & \textbf{0.730} & \textbf{0.946} \\ \bottomrule
\end{tabular}
\label{tab:compare_ip2p_no_ip2p_progress}
\vspace{-0.3cm}
\end{table}

\subsection{Real-Robot Experiments}

\begin{figure*}[!t]
\centering
\includegraphics[width=0.9\textwidth]{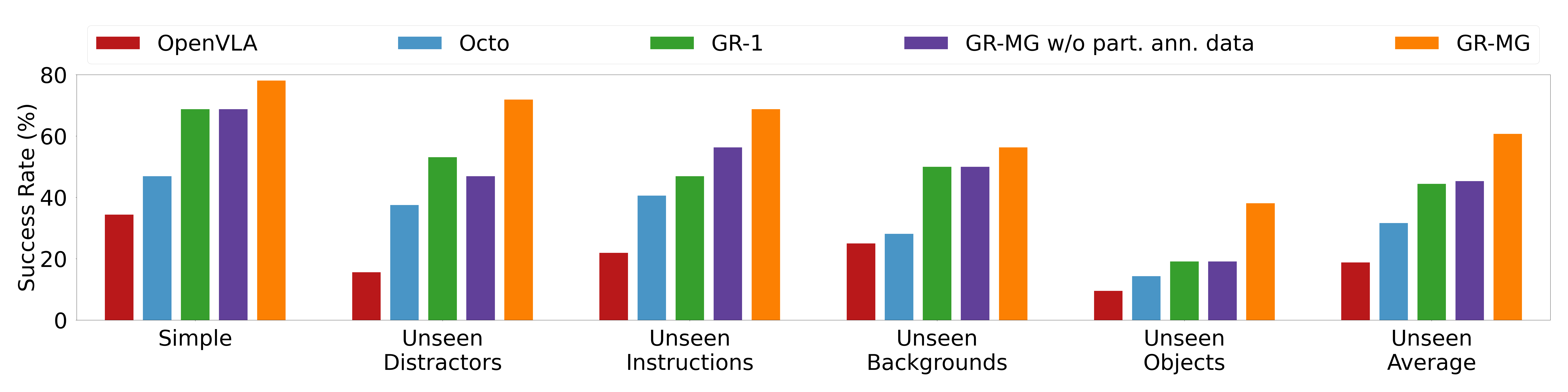}
\caption{
\textbf{Success Rates of Real-Robot Experiments.}
Unseen Average shows the average success rate of the four unseen generalization settings.
}
\label{fig:real_result}
\vspace{-0.4cm}
\end{figure*}

\subsubsection{Experiment Settings}
To evaluate the performance of GR-MG in the real world, we perform experiments on a real robot platform.
The platform consists of a Kinova Gen-3 robot arm equipped with a Robotiq 2F-85 parallel-jaw gripper and two cameras, \textit{i.e.}, one static camera for capturing the workspace and another camera mounted on the end-effector.
The training dataset consists of 18k human demonstrations across 37 tasks, which include 23 pick-and-place tasks and 14 non pick-and-place tasks such as pouring, flipping, and rotating. 
See Fig.~\ref{fig:data} for some examples.

In this experiment, we aim to verify the effectiveness of leveraging data w/o action labels in training.
Specifically, we combine the data from Something-Something-V2~\cite{goyal2017something} and RT-1~\cite{brohan2022rt} with our real robot data to create the training dataset for training the goal image generation model.
It is worth noting that any videos with text annotations can be included in the training process.

We design five different settings to evaluate the model performance: Simple, Unseen Distractors, Unseen Instructions, Unseen Backgrounds, and Unseen Objects.
In Simple, the scene is set to be similar to those in the training data.
In Unseen Distractors, unseen distractors are added to the scene.
In Unseen Instructions, we follow~\cite{li2023vision} and use GPT-4 to generate unseen synonyms for the verbs in the instructions.
For example, we replace "pick up" with "take", "cap" with "cover", and "stack" with "pile".
In Unseen Backgrounds, the background is modified by introducing two new tablecloths that were not present in the training data.
In Unseen Objects, the robot is instructed to manipulate objects that were not included in the training dataset. 
And the language instructions are adjusted accordingly, \textit{i.e.}, the language instructions are also unseen.
In total, we evaluate 58 different tasks: 37 of which were seen during training, while the rest were unseen.
See the appendix on the project page for the full list of training tasks and the 58 evaluated tasks.
We compare the performance of GR-MG with OpenVLA~\cite{kim2024openvla}, Octo~\cite{team2024octo}, and GR-1~\cite{wu2023unleashing}.
To study the effectiveness of leveraging data w/o action labels, we also compare with the variant which only uses fully-annotated data for training the goal image generation model, GR-MG w/o part. ann. data.

\subsubsection{{Results}}
Results are shown in Fig.~\ref{fig:real_result}.
Real-robot rollout videos can be found in the supplementary material and on the project page.
GR-MG outperforms all the baseline methods in all five evaluation settings.
A typical failure mode of OpenVLA is inaccurate grasping.
We believe this is because it uses an discretized action space, making the action prediction inaccurate.
Also, it does not support inputs of observation history and wrist camera observation, making it difficult to accurately identify the appropriate timestep to close/open the gripper in some tasks.
Octo supports inputs of observation history, proprioceptive data, and observation from the wrist camera.
However, it is not able to generalize well in unseen settings, especially for Unseen Backgrounds and Unseen Objects.
GR-1 performs well in most settings, but it is not able to pick the correct object in the setting of Unseen Objects.
GR-MG is able to generalize well to unseen settings.
The performance degradation in Unseen Instructions is modest for GR-MG compared to other baseline methods.
We believe the powerful generalization capabilities of GR-MG stem from the video pre-training for both the goal image generation model and the policy.
It allows the model to harness the prior knowledge gained from pre-training to enhance its performance in downstream manipulation tasks.

By comparing with GR-MG w/o part. ann. data, we conclude that additional partially-annotated data w/o action labels are crucial for enhancing the generalization capabilities of GR-MG. 
We qualitatively compare the goal images generated by GR-MG and GR-MG w/o part. ann. data in Fig.~\ref{fig:compare_ip2p}(b).
GR-MG generates high-quality and accurate goal images, while the goal images generated by GR-MG w/o part. ann. data do not adhere to the language instructions.
In the setting of Unseen Objects (second row of Fig.~\ref{fig:compare_ip2p}(b)), the absence of additional partially-annotated training data increases the likelihood of hallucination: objects from the training set will unexpectedly appear in the generated goal images even if they are not present in the current scene.
These findings indicate that incorporating more data w/o action labels into training enhances 1) understanding of language semantics and 2) robustness against out-of-distribution data.

\subsubsection{Few-Shot Learning of Novel Skills}

We further perform experiments to evaluate the few-shot learning capabilities of GR-MG.
In particular, we first hold out 8 tasks from the 37 tasks and train the model on the rest 29 tasks.
The total number of trajectories used for training is 15k.
These 8 tasks include 7 novel skills which are unseen in the 29 training tasks.
For example, we hold out "wipe the cutting board" and there are no wiping tasks in the 29 training tasks.
For the full list of all the held-out tasks, please see the appendix on the project page.
After training, we further finetune the model with data containing 10 (or 30) trajectories per held-out task.
That is, the total number of trajectories used for finetuning in this stage is 80 (or 240).
We evaluate on the 8 held-out tasks and results are shown in Tab.~\ref{tab:few-shot}.
GR-MG outperforms all the comparing baseline methods in both 10-shot and 30-shot settings.
We observe that the main bottleneck of GR-MG lies within the policy.
The goal image generation model is able to generate accurate goal images on the these novel skills after few-shot finetuning, but the policy struggles to perform well.
To tackle this challenge, we plan to investigate scaling up the training of the policy with more real-world data w/o text labels in the future.

\vspace{-0.1cm}

{
\setlength{\tabcolsep}{18pt}
\renewcommand{\arraystretch}{0.95}
\begin{table}[]
\centering
\caption{Success Rates of Few-Shot Learning}
\begin{tabular}{@{}clcllcll@{}}
\toprule
{Methods} & {10-shot} & {30-shot} \\ 
\midrule
{OpenVLA~\cite{kim2024openvla}} & {0.0\%} & {2.5\%} \\
{Octo~\cite{team2024octo}} & {0.0\%} & {0.0\%} \\
{GR-1~\cite{wu2023unleashing}} & {2.5\%} & {22.5\%} \\
{GR-MG w/o part. ann. data} & {10.0\%} & {27.5\%} \\
{GR-MG (Ours)} & \textbf{17.5\%} & \textbf{37.5\%} \\ 
\bottomrule
\end{tabular}
\label{tab:few-shot}
\vspace{-0.3cm}
\end{table}
}

\section{Conclusions}
\label{sec:conclusions}
In this paper, we introduce GR-MG, a novel method that leverages multi-modal goals to predict actions.
GR-MG uses both a language and a goal image to condition the action prediction.
It requires only the language as input and generate the goal image based on the language via a goal image generation model. 
This design enables GR-MG to effectively leverage large amounts of partially-annotated data that are missing either text or action labels. 
To improve the robustness of the generated goal images, we incorporate task progress information in the goal image generation model. 
GR-MG showcases exceptional performance in both simulation and real-world experiments, showing strong generalization capabilities across various out-of-distribution settings.
It can also efficiently learn novel skills in a few-shot setting.
We plan to scale up the training of both the goal image generation model and the policy by incorporating more partially-annotated data in the future.
Additionally, we plan to investigate integrating depth information to further improve the accuracy of action prediction.

\section{ACKNOWLEDGEMENT}
\label{sec:acknowledgment}
This work was partially supported by National Natural Science Foundation of China (62322607, 62236010 and 62276261), and Youth Innovation Promotion Association of Chinese Academy of Sciences under Grant 2021128.


\bibliographystyle{IEEEtran}
\bibliography{references}

\end{document}